\def\year{2018}\relax
\documentclass[letterpaper]{article} %DO NOT CHANGE THIS
\usepackage{aaai18}  %Required
\usepackage{times}  %Required
\usepackage{helvet}  %Required
\usepackage{courier}  %Required
\usepackage{url}  %Required
\usepackage{graphicx}  %Required
\usepackage{amsmath}
\usepackage{amsthm}
\usepackage{amsfonts}       % blackboard math symbols
\usepackage{nicefrac}       % compact symbols for 1/2, etc.
\usepackage{microtype}      % microtypography
\usepackage{subcaption}
\usepackage{algorithm}		% For algorithms	
\usepackage{algorithmic}	% For algorithmsx
\title{Multi-Step Reinforcement Learning: A Unifying Algorithm}
\author{Kristopher De Asis,$^1$ J. Fernando Hernandez-Garcia,$^1$ G. Zacharias Holland,$^1$ Richard S. Sutton\\
Reinforcement Learning and Artificial Intelligence Laboratory, University of Alberta \\
\{kldeasis,jfhernan,gholland,rsutton\}@ualberta.ca
}

\begin{document}
\maketitle
\begin{abstract}
Unifying seemingly disparate algorithmic ideas to produce better performing algorithms has been a longstanding goal in reinforcement learning.
As a primary example, TD($\lambda$) elegantly unifies one-step TD prediction with Monte Carlo methods through the use of eligibility traces and the trace-decay parameter $\lambda$.
Currently, there are a multitude of algorithms that can be used to perform TD control, including Sarsa, $Q$-learning, and Expected Sarsa.
These methods are often studied in the one-step case, but they can be extended across multiple time steps to achieve better performance.
Each of these algorithms is seemingly distinct, and no one dominates the others for all problems.
In this paper, we study a new multi-step action-value algorithm called $Q(\sigma)$ that unifies and generalizes these existing algorithms, while subsuming them as special cases.
A new parameter, $\sigma$, is introduced to allow the degree of sampling performed by the algorithm at each step during its backup to be continuously varied, 
with Sarsa existing at one extreme (full sampling), and Expected Sarsa existing at the other (pure expectation).
$Q(\sigma)$ is generally applicable to both on- and off-policy learning, but in this work we focus on experiments in the on-policy case.
Our results show that an intermediate value of $\sigma$, which results in a mixture of the existing algorithms, performs better than either extreme.
The mixture can also be varied dynamically which can result in even greater performance.
\end{abstract}

\newtheorem{theorem}{Theorem}
\section{The Landscape of TD Algorithms}
\label{sec:introduction}
\footnotetext[1]{Authors contributed equally, and are listed alphabetically.}
\textit{Temporal-difference} (TD) methods~\cite{sutton98} are an important concept in \textit{reinforcement learning} (RL) that combines ideas from Monte Carlo and dynamic programming methods.
TD methods allow learning to occur directly from raw experience in the absence of a model of the environment's dynamics, like with Monte Carlo methods, while also allowing estimates to be updated based on other learned estimates without waiting for a final result, like with dynamic programming.
The core concepts of TD methods provide a flexible framework for creating a variety of powerful algorithms that can be used for both prediction and control.

There are a number of TD control methods that have been proposed.
$Q$-learning~\cite{watkins1989qlearn,watkins1992q} is arguably the most popular, and is considered an \textit{off-policy} method because the policy generating the behaviour (the \textit{behaviour policy}), and the policy that is being learned (the \textit{target policy}) are different.
Sarsa~\cite{rummery1994,sutton1996} is the classical \textit{on-policy} control method, where the behaviour and target policies are the same.
However, Sarsa can be extended to learn off-policy with the use of importance sampling~\cite{precup2000}.
Expected Sarsa is an extension of Sarsa that, instead of using the action-value of the next state to update the value of the current state, uses the expectation of all the subsequent action-values of the current state with respect to the target policy.
Expected Sarsa has been studied as a strictly on-policy method~\cite{vanSeijen2009}, 
but in this paper we present a more general version that can be used for both on- and off-policy learning and that also subsumes $Q$-learning.
All of these methods are often described in the simple one-step case, 
but they can also be extended across multiple time steps.

The TD($\lambda$) algorithm unifies one-step TD learning with Monte Carlo methods~\cite{sutton1988learning}. 
Through the use of eligibility traces, and the trace-decay parameter, $\lambda\in [0,1]$, a spectrum of algorithms is created.
At one end, $\lambda=1$, exists Monte Carlo methods, and at the other, $\lambda=0$, exists one-step TD learning.
In the middle of the spectrum are intermediate methods which can perform better than the methods at either extreme~\cite{sutton98}.
The concept of eligibility traces can also be applied to TD control methods such as Sarsa and $Q$-learning, which can create more efficient learning and produce better performance~\cite{rummery1995}.

Multi-step TD methods are usually thought of in terms of an average of many multi-step returns of differing lengths and are often associated with eligibility traces, as is the case with TD($\lambda$).
However, it is also natural to think of them in terms of individual $n$-step returns with their associated $n$-step backup~\cite{sutton98}.
We refer to each of these individual backups as \textit{atomic backups}, whereas the combination of several atomic backups of different lengths creates a \textit{compound backup}.

In the existing literature, it is not clear how best to extend one-step Expected Sarsa to a multi-step algorithm.
The \textit{Tree-backup} algorithm was originally presented as a method to perform off-policy evaluation when the behaviour policy is non-Markov, non-stationary or completely unknown~\cite{precup2000}.
In this paper, we re-present Tree-backup as a natural multi-step extension of Expected Sarsa.
Instead of performing the updates with entirely sampled transitions as with multi-step Sarsa, Tree-backup performs the update using the expected values of all the actions at each transition.

$Q(\sigma)$ is an algorithm that was first proposed by Sutton and Barto~(\citeyear{sutton2017}) which unifies and generalizes the existing multi-step TD control methods.
The degree of sampling performed by the algorithm is controlled by the sampling parameter, $\sigma$.
At one extreme ($\sigma=1$) exists Sarsa (full sampling), and at the other ($\sigma=0$) exists Tree-backup (pure expectation).
Intermediate values of $\sigma$ create algorithms with a mixture of sampling and expectation,
and $\sigma$ can be interpreted as a way to control the bias-variance trade-off inherent in multi-step TD algorithms.

In this work, on problems with a tabular representation and a problem requiring function approximation, we show that an intermediate value of $\sigma$ can outperform the algorithms that exist at either extreme.
In addition, we show that $\sigma$ can be varied dynamically to produce even greater performance.
We limit our discussion of $Q(\sigma)$ to the atomic multi-step case without eligibility traces, but a natural extension is to make use of compound backups and is an avenue for future research.
Furthermore, $Q(\sigma)$ is generally applicable to both on- and off-policy learning, but for our initial empirical study we examined only on-policy prediction and control problems.

\section{MDPs and One-step Solution Methods}
\label{sec:background}
% MDP definition
The sequential decision problem encountered in RL is often modeled as a \textit{Markov decision process} (MDP). Under this framework, an \textit{agent} and the environment interact over a sequence of discrete time steps $t$. At every time step, the agent receives information about the environment's \textit{state}, $S_t \in \mathcal{S}$, where $\mathcal{S}$ is the set of all possible states. The agent uses this information to select an \textit{action}, $A_t$, from the set of all possible actions $\mathcal{A}$. Based on the behavior of the agent and the state of the environment, the agent receives a \textit{reward}, $R_{t+1} \in \mathbb{R}$, and moves to another state, $S_{t+1} \in \mathcal{S}$, with a \textit{state-transition probability} $p(s'|s,a)={P(S_{t+1}=s'|S_t=s,A_t=a)}$, for $a\in \mathcal{A}$ and $s,s' \in \mathcal{S}$.

% Introduction of policy and policy iteration
The agent behaves according to a \textit{policy} $\pi(a|s)$, which is a probability distribution over the set $\mathcal{S}\times\mathcal{A}$. Through the process of policy iteration~\cite{sutton98}, the agent learns the optimal policy, $\pi^*$, that maximizes the expected discounted return:
\begin{equation}
	G_t = R_{t+1} + \gamma R_{t+2} + \gamma^2 R_{t+3} + ... = \sum_{k=0}^{T-t-1} \gamma^k R_{t+1+k},
\end{equation}
for a discount factor $\gamma \in [0,1)$ and $T = \infty$ for continuing tasks, or $\gamma \in [0,1]$ and $T$ equal to the final time step in episodic tasks.
% for a discount factor $\gamma \in [0,1]$ and $T$ equal to infinity for continuing tasks, or the final time step in episodic tasks.

\begin{figure}
	\includegraphics[width=\linewidth]{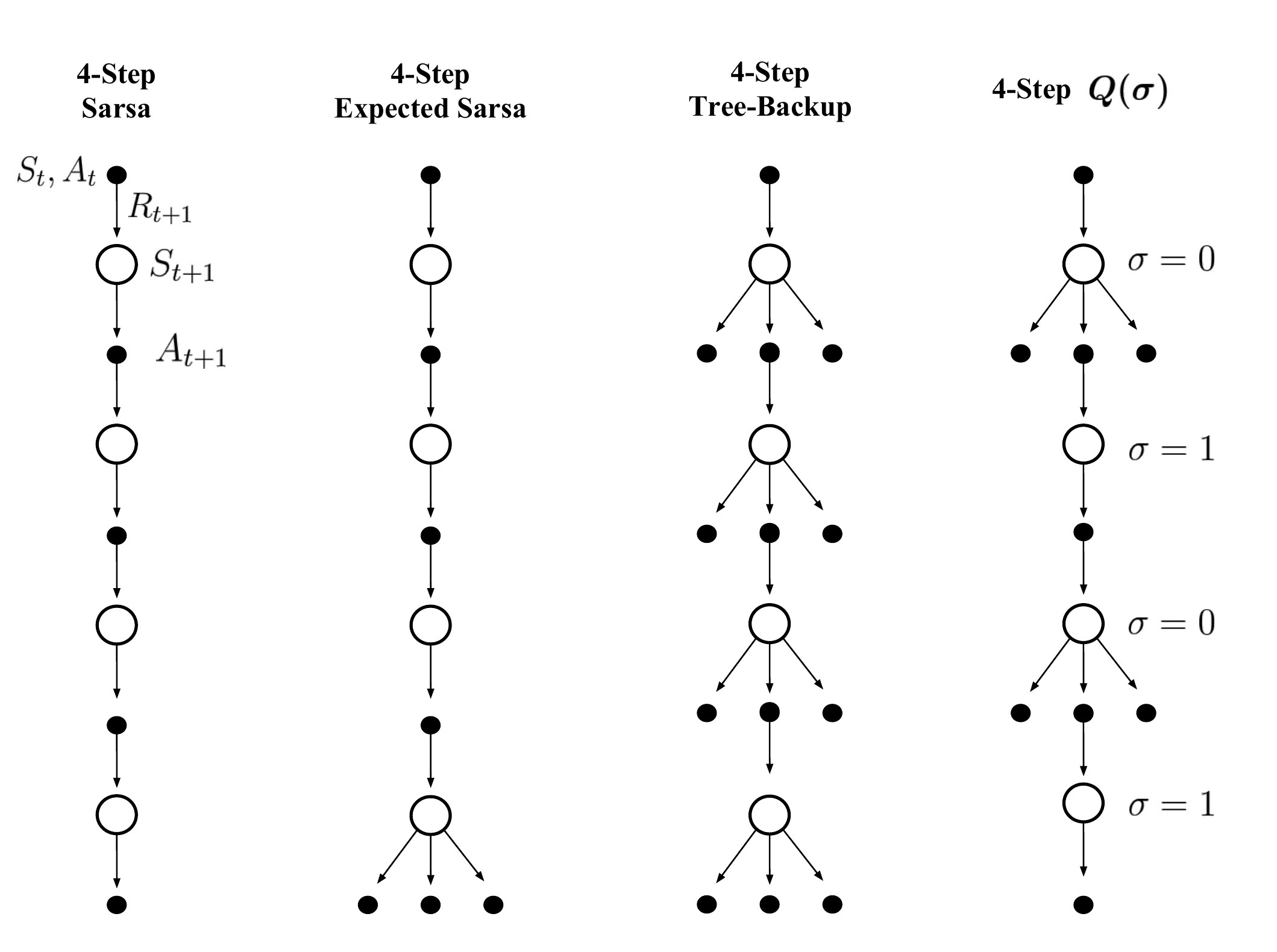}
	\caption{Backup diagrams for atomic 4-step Sarsa, Expected Sarsa, Tree-backup, and $Q(\sigma)$. Here we can see that $Q(\sigma)$ encompasses the other three algorithms based on the setting of $\sigma$.}
    \label{fig:backupdiag}
\end{figure}

% state value functions and action value functions
TD algorithms strive to maximize the expected return by computing value functions that estimate the expected future rewards in terms of the elements of the environment and the actions of the agent. The \textit{state-value function} is the expected return when the agent is in a state $s$ and follows policy $\pi$, defined as $v_\pi(s)=\mathbb{E}_\pi[G_t|S_t=s]$. For control, most of the time we focus on estimating the \textit{action-value function}, which is the expected return when the agent takes an action $a$, in a state $s$, while following a policy $\pi$, and is defined as:
\begin{equation}
	q_\pi(s,a)=\mathbb{E}_\pi[G_t|S_t=s,A_t=a].
\label{eqn:avFunct}
\end{equation}
Equation \ref{eqn:avFunct} can be estimated iteratively by observing new rewards, bootstrapping on old estimates of $q_\pi$, and using the update rule:
% update rule
\begin{align}
Q(S_t,A_t) & \leftarrow \ Q(S_t,A_t)  \label{eqn:updtrule} \\ 
		   & + \alpha[R_{t+1} + \gamma Q(S_{t+1}, A_{t+1}) - Q(S_t,A_t)], 				                 			   			  \nonumber
\end{align}
where $\alpha \in (0,1]$ is the step size parameter. Update rules are also known as \textit{backup} operations because they transfer information back from future states to the current one.
A common way to visualize backup operations is by using backup diagrams such as the ones depicted in Figure \ref{fig:backupdiag}.

For clarity, the algorithmic ideas in this paper are presented initially as \textit{tabular solution methods}, but we also extend them to use function approximation, and thus they also serve as \textit{approximate solution methods}. 
% More details about the algorithms as approximate solution methods are available in the appendix.

% \textit{Tabular solution methods} can be used when the state and action spaces are small enough so that it is possible to maintain the estimates of the value functions in an array or table.
% When the state space is large and/or continuous, \textit{approximate solution methods} need to be used, which combine RL algorithms with some kind of function approximation scheme.
% For simplicity, we present the algorithmic ideas in this paper as tabular solution methods, 
% but they can also be extended to use function approximation, and thus can serve also as approximate solution methods.\footnote{See the appendix for a presentation of the algorithms for use with function approximation.}
% In our experiments we study problems that require tabular solution methods and a problem that requires an approximate solution method using function approximation.

% introduction of Sarsa and TD error
The term in brackets in (\ref{eqn:updtrule}):
\begin{align}
	\delta_t^{S} = R_{t+1} + \gamma Q(S_{t+1}, A_{t+1}) - Q(S_t,A_t),
    \label{eqn:sarsaTDerror}
\end{align}
is also known as the \textit{TD error}, denoted $\delta_t$. TD control methods are characterized by their TD error; for example, the TD error in (\ref{eqn:sarsaTDerror}) corresponds to the classic on-policy method known as Sarsa.

% introduction to off-policy learning
Because learning requires a certain amount of exploration, behaving greedily with respect to the estimated optimal policy is often infeasible. Therefore, agents are often trained under $\epsilon$-greedy policies for which the agent only chooses the optimal action with a probability $(1-\epsilon)$ and behaves randomly with probability $\epsilon$, for $\epsilon \in [0,1]$. Nevertheless, learning the optimal policy is possible if it is done off-policy. When the agent is learning off-policy, it behaves according to a behavior policy, $\mu$, while learning a target policy, $\pi$. 
% introduction of Expected Sarsa
This can be achieved by using another TD control method, Expected Sarsa. In contrast with Sarsa, Expected Sarsa behaves according to the behavior policy, but updates its estimate by taking an expectation of $Q(S_t,A_t)$ over the actions at time $t$, according to the target policy~\cite{vanSeijen2009}. For convenience, let the expected action-value be defined as:
\begin{equation}
	V_{t+1} = \sum_a \pi(a|S_{t+1})Q(S_{t+1},a). 
    \label{eqn:expectedActionValue}
\end{equation}
Then, the TD error of Expected Sarsa can be written as:
\begin{equation}
	\delta_t^{ES} = R_{t+1} + \gamma V_{t+1} - Q(S_t,A_t).
	\label{eqn:expectedSarsa}
\end{equation}

% Q-learning
A special case of Expected Sarsa is $Q$-learning, where the estimate is updated according to the maximum of $Q(S_t,a)$ over the actions~\cite{watkins1989qlearn}:
\begin{align}
	\delta_t^{QL} = R_{t+1} + \gamma \max_a Q(S_{t+1}, a) - Q(S_t,A_t).
\end{align}
$Q$-learning is the resulting algorithm when the target policy of Expected Sarsa is the greedy policy with respect to $Q$. 

%%% Fig: Backup Diagrams %%%

\section{Atomic Multi-Step Algorithms}
The TD methods presented in the previous section can be generalized even further by bootstrapping over longer time intervals. This has been shown to decrease the bias of the update at the cost of increasing the variance~\cite{jaakkola1994}. 
Nevertheless, in many cases it is possible to achieve better performance by choosing a value for the backup length parameter, $n$, greater than one~\cite{sutton98}.
%which creates an atomic multi-step algorithm 
We refer to algorithms which make use of a multi-step atomic backup as \textit{atomic multi-step algorithms}.
%\footnote{Sutton and Barto (1998) refer to these as \textit{simple} multi-step algorithms.}
Just like how one-step methods are defined by their TD error, each atomic multi-step algorithm is characterized by its $n$-step return. For atomic multi-step Sarsa, the $n$-step return is:
\begin{align}
	G_{t:t+n} =& R_{t+1} + \gamma R_{t+2} + \gamma^2 R_{t+3} + ... \nonumber \\
    &+ \gamma^{n-1} R_{t+n} 
    		  + \gamma^n Q_{t+n-1}(S_{t+n},A_{t+n}),  \nonumber \\
    		  =& \sum_{k=0}^{n-1} \gamma^k R_{t+k+1} + \gamma^n Q_{t+n-1}(S_{t+n}, A_{t+n}),
              \label{eqn:nstepSarsa}
\end{align}
where $Q_{t+n-1}$ is the estimate of $q_{\pi}$ at time $t+n-1$, and the subscript range, $t:t+n$, denotes the length of the backup. $n$-step Sarsa can be adapted for off-policy learning by introducing an importance sampling ratio term~\cite{precup2000}:
\begin{equation}
	\rho_t^{t+n} = \prod_{k=t}^{\tau} \frac{\pi(A_k|S_k)}{\mu(A_k|S_k)},
    \label{eqn:rho}
\end{equation}
and multiplying it with the TD error to get the following update rule:
\begin{align}
	Q_{t+n}(S_t,A_t) & \leftarrow  Q_{t+n-1}(S_t,A_t) \label{eqn:impSampupdate} \\
  	                 & + \alpha \rho_{t+1}^{t+n}[G_{t:t+n} - Q_{t+n-1}(S_t,A_t)], \nonumber 
\end{align}
where $\tau = \min(t+n-1, T-1)$ is the time step before the end of the update or before the end of the episode. In the update, the action-values for all other states remain the same -- 
%i.e. $Q_{t+n}(s,a)=Q_{t+n-1}(s,a),\forall\ s,a \in \{s',a' | a' \neq A_t, s' \neq S_t\}$.
% I don't think the statement above was specific enough cause it didn't include cases where a != A_t, but s = S_t
i.e. $Q_{t+n}(s,a)=Q_{t+n-1}(s,a),\forall\ s \neq S_t$, and $a \neq A_t$.
This update rule is not only applicable for off-policy $n$-step Sarsa, but is a generally useful form for other atomic multi-step algorithms as well.
We present the algorithms in this work as general off-policy solution methods, but in the experiments section we evaluate them empirically on-policy only which provides useful insight into their behaviour. 
We defer the empirical study and comparison of the algorithms in an off-policy setting to future work.

% n-step expected sarsa
Expected Sarsa can also be generalized to a multi-step method by using the return:
\begin{equation}
	G_{t:t+n} = R_{t+1} + \gamma R_{t+2} + \gamma^2 R_{t+3} + ... + \gamma^n V_{t+n}.
    \label{eqn:nstepESarsa}
\end{equation}
% n-step update rule

The first $n-1$ states and actions are sampled according to the behaviour policy, as with $n$-step Sarsa, but the last state is backed up according to the expected action-value under the target policy.
To make $n$-step Expected Sarsa entirely off-policy, an importance sampling ratio term can also be introduced, but it needs to omit the last time step.
The resulting update would be the same as in (\ref{eqn:impSampupdate}), but would use $\rho^{t+n-1}_{t+1}$ and the $n$-step return for $n$-step Expected Sarsa from (\ref{eqn:nstepESarsa}).

A drawback to using importance sampling to learn off-policy is that it can create high variance which must be compensated for by using small step sizes; 
this can slow learning~\cite{precup2000}.
In the next section we present a method that is also a generalization of Expected Sarsa, but that can learn off-policy without importance sampling.

%%%%%%%%%%%%%%%%%%%%%%%%%%%%%%%%%%% Tree-backup %%%%%%%%%%%%%%%%%%%%%%%%%%%%%%%%%%%%%%%
\section{Tree-backup}
As shown in (\ref{eqn:nstepESarsa}), the TD return of $n$-step Expected Sarsa is calculated by taking an expectation over the actions at the last step of the backup. However, it is possible to extend this idea to every time step of the backup by taking an expectation at every step~\cite{precup2000}. The resulting algorithm is a multi-step generalization of Expected Sarsa that is known as Tree-backup because of its characteristic backup diagram (Figure \ref{fig:backupdiag}).
% Added to address Reviewer's 3 Comments  (Second item in the check list)
Moreover, just like Expected Sarsa and Q-learning, this proposed generalization does not require importance sampling to be applied off-policy. Hence, it could be argued that it is a more appropriate generalization of Expected Sarsa to multi-step learning \cite{sutton2017}. Because Expected Sarsa subsumes $Q$-learning, Tree-backup can also be thought of as a multi-step generalization of $Q$-learning if the target policy is greedy with respect to the action-value function.

Tree-backup has several advantages over $n$-step Expected Sarsa.
Tree-backup has the capacity for learning off-policy without the need for importance sampling, reducing the variance due to the importance sampling ratios.
Additionally, because an importance sampling ratio does not need to be computed, the behavior policy does not need to be stationary, Markov, or even known~\cite{precup2000}.

Each branch of the tree represents an action, while the main branch represents the action taken at time $t$. The value of each of the branches is the value of $Q_{t+n}(S_t,A_t)$ for the corresponding $t$, whereas the value of each segment of the main branch is the reward at the corresponding time step. The $n$-step return is the sum of the values of each branch weighted by the product of the probabilities of the actions leading to the branch and multiplied by the corresponding power of the discount term. For clarity, it is easier to present the $n$-step return of the Tree-backup algorithm in terms of the TD error of Expected Sarsa from (\ref{eqn:expectedSarsa}):
\begin{align}
	G_{t:t+n}  =& Q_{t-1}(S_t,A_t) 
    		  + \sum_{k=t}^{\tau} \delta_k^{ES} 															  \prod_{i=t+1}^{k} \gamma \pi(A_i|S_i).
\label{eqn:treeBackup} 
\end{align}
% For a full derivation of this return, see the appendix. 
This atomic version of multi-step Tree-backup was first presented by Sutton and Barto (\citeyear{sutton2017}).

% KD: note about TB's bias here
As a result of the product term in (\ref{eqn:treeBackup}), 
%it can be seen that %
in addition to the discount factor $\gamma$, future rewards are further discounted by the probabilities of the actions taken. The Tree-backup algorithm therefore assigns less weight to the reward sequence received, and compensates by bootstrapping off of the values of actions not taken. Due to this, Tree-backup is more biased than Sarsa in the multi-step case with a stochastic policy, as Sarsa gives full weight to every reward received prior to bootstrapping. However, this increase in bias (towards the estimates in the value function) is traded off with decreased variance in the reward sequence from taking expectations. 
% A more detailed bias analysis of the $n$-step returns is presented in the appendix.

%%%%%%%%%%%%%%%%%%%%%%%%%%%%%%%%%%%%%%%%% Q-Sigma $$$$$%%%%%%%%%%%%%%%%%%%%%%%%%%%%%%%%%%%
\section{The $\boldsymbol{Q(\sigma)}$ Algorithm}
\label{sec:qsigma}
In the previous sections we have incrementally introduced several generalizations for the TD control methods Sarsa and Expected Sarsa, and in this section we present an algorithm that unifies them called $Q(\sigma)$.

Sarsa can be generalized to an atomic multi-step algorithm by using an $n$-step return, and $n$-step Sarsa generalizes to an off-policy algorithm through the use of importance sampling.
In contrast, Expected Sarsa can learn off-policy without the need for importance sampling, 
and generalizes to the atomic multi-step algorithms: Tree-backup and $n$-step Expected Sarsa. 
All of the algorithms presented so far can be broadly categorized into two families:
those that backup their actions as samples, like Sarsa; and those that consider an expectation over all actions in their backup, like Expected Sarsa and Tree-backup.
%\footnote{It may be useful to note that $n$-step Expected Sarsa belongs to both families since it samples every action, except the last one, for which it considers the expectation.}
In this section, we introduce a method to unify both families of algorithms by introducing a new parameter, $\sigma$. 
The possibility of unifying Sarsa and Tree-backup was first suggested by Precup et al. (\citeyear{precup2000}), and the first formulation of $Q(\sigma)$ was presented by Sutton and Barto~(\citeyear{sutton2017}).

The intuition behind $Q(\sigma)$ is based on the idea that we have a choice to update the estimate of $q_\pi$ based on one action sampled from the set of possible future actions, or based on the expectation over the possible future actions.
For example, with $n$-step Sarsa, a sample is taken at every step of the backup, whereas with the Tree-backup algorithm, an expectation is taken instead. 
However, the choice of sampling or expectation does not have to remain constant for every step of the backup. Furthermore, the backup at a time step $t$ could be based on a weighted average of both sampling and expectation. In order to implement this, the parameter, $\sigma_t \in [0,1]$, is introduced to control the degree of sampling at each step of the backup. 
Thus, the TD error of $Q(\sigma)$ can be represented in terms of a weighted sum of the TD errors of Sarsa and Expected Sarsa:
\begin{align}
	\delta_t^{\sigma} =& \ \sigma_{t+1} \delta_t^{S} + (1-\sigma_{t+1}) \delta_t^{ES},  \nonumber \\
    =& \ R_{t+1} + \gamma[\sigma_{t+1} Q_{t}(S_{t+1},A_{t+1}) + (1-\sigma_{t+1}) V_{t+1}] \nonumber \\ 
    &- Q_{t-1}(S_t,A_t).
\end{align}
The $n$-step return is then:
\begin{align}
	G_{t:t+n} =& \ Q_{t-1}(S_t,A_t) \label{eqn:qsigmaNStep} \\
    &+ \sum_{k=t}^{\tau} \delta_k^\sigma 
    \prod_{i=t+1}^k \gamma[(1-\sigma_i)\pi(A_i|S_i) + \sigma_i]. \nonumber
\end{align}
% For a full derivation of this return, see the appendix.
Moreover, the importance sampling ratio from (\ref{eqn:rho}) can be modified to include $\sigma$ as follows:
\begin{align}
	\rho_{t+1}^{t+n} = 	\prod_{k=t+1}^{\tau} \bigg(\sigma_k 													\frac{\pi(A_k|S_k)}{\mu(A_k|S_k)} + 1 - \sigma_k\bigg).
    				\label{eqn:qsigmaRho}
\end{align}
The update rule for $Q(\sigma$) can then be obtained by using $G_{t:t+n}$ from (\ref{eqn:qsigmaNStep}) and $\rho_{t+1}^{t+n}$ from (\ref{eqn:qsigmaRho}), with the update rule from (\ref{eqn:impSampupdate}).
Algorithm \ref{alg:qsigma} shows the pseudocode for the complete off-policy $n$-step $Q(\sigma)$ algorithm. 

\begin{algorithm}[tb]
	\caption{Off-policy $n$-step $Q(\sigma)$ for estimating $q_\pi$}
	\label{alg:qsigma}
	\begin{algorithmic}
    	\STATE Input: a behaviour policy $\mu$ and a target policy $\pi$
    	\STATE Initialize $S_0 \neq$ terminal; select $A_0$ according to $\mu(.|S_0)$
        \STATE Store $S_0$, $A_0$, and $Q(S_0,A_0)$
        \FOR {$t=0,1,2,...,T+n-1$}
        	\IF{$t < T$}
            	\STATE Take action $A_t$; observe $R$ and $S_{t+1}$
                \STATE Store $S_{t+1}$
                \IF{$S_{t+1}$ is terminal}
                    \STATE Store: $\delta_t^\sigma = R - Q(S_t, A_t)$
                \ELSE
                	\STATE Select $A_{t+1}$ according to $\mu(\cdot|S_{t+1})$ and Store
                    \STATE Store: $Q(S_{t+1}, A_{t+1})$, $\sigma_{t+1}$, $\pi(A_{t+1}|S_{t+1})$
                    \STATE Store: $\delta_t^\sigma = R +\gamma[\sigma_{t+1} Q(S_{t+1}, A_{t+1})$
                    \STATE $ \qquad \qquad + (1-\sigma_{t+1})V_{t+1}] - Q(S_t,A_t)$
                    \STATE Store: $\rho_{t+1}=\frac{\pi(A_{t+1}|S_{t+1})}{\mu(S_{t+1}|A_{t+1})}$
                \ENDIF
            \ENDIF
            \STATE $\tau\leftarrow t-n+1$
            \IF{$\tau\geq 0$}
            	\STATE $\rho \leftarrow 1$
                \STATE $E \leftarrow 1$
                \STATE $G \leftarrow Q(S_\tau,A_\tau)$
                \FOR {$k=\tau,...,\min(\tau+n-1,T-1)$}
                	\STATE $G \leftarrow G + E\delta^\sigma_k$
                    \STATE $E \leftarrow \gamma E\left[(1-\sigma_k)\pi(A_{k+1}|S_{k+1})+\sigma_{k+1}\right]$
                    \STATE $\rho \leftarrow \rho(1-\sigma_k+\sigma_k\rho_k)$
                \ENDFOR
                \STATE $Q(S_\tau,A_\tau) \leftarrow Q(S_\tau,A_\tau) + \alpha \rho [G - Q(S_\tau,A_\tau)]$
            \ENDIF
        \ENDFOR
	\end{algorithmic}
\end{algorithm}

Additionally, a proof for one-step $Q(\sigma)$ is readily available by applying the results from Jakkola et al. (\citeyear{jaakkola1994}), Singh et al. (\citeyear{Singh2000}), and van Seijen et al. (\citeyear{vanSeijen2009}).

\begin{theorem}
\label{thm:qsigma}
The one-step $Q(\sigma$) estimate defined by 
%% Q(sigma) update
\begin{align}
& Q_{t+1}(S_t,A_t) = (1-\alpha_t) Q_t(S_t,A_t) + \alpha_t [R_{t+1} \nonumber \\
& \qquad + \gamma (\sigma_{t+1} Q_{t+1}(S_{t+1},A_{t+1}) + (1-\sigma_{t+1})V_{t+1})],
\label{eqn:1stepqs}
\end{align} 
converges to the optimal policy when the following conditions are satisfied:
\begin{enumerate}
\item The size of the set $\mathcal{S} \times \mathcal{A}$ is finite.
\item $\alpha_t = \alpha_t(S_t,A_t) \in [0,1]$, $\sum_t \alpha_t = \infty$, $\sum_t \alpha_t^2 < \infty$ w.p. 1 and $\forall (s,a) \neq (S_t,A_t): \alpha_t(S_t,A_t) = 0$.
\item The policy is greedy in the limit with infinite exploration.
\item The reward function is bounded.
\end{enumerate}
\end{theorem}

We defer the full details of the proof to the appendix; however, 
There are two important results from the proof that are worth emphasizing. First, just as with one-step $Q$-learning, Sarsa, and Expected Sarsa, one-step $Q(\sigma)$ can be used to learn optimal action-value functions. Second, at each time step $t$ it is possible to choose a $\sigma_t$ such that the contraction property of the $Q(\sigma)$ update is less than or equal to the contraction induced by the Sarsa or Expected Sarsa updates. This implies that it is possible to choose $\sigma_t$ at every time step in order to speed up convergence. 

It is important to note that every TD control method presented thus far can be obtained with $Q(\sigma)$ by varying the sampling parameter, $\sigma$;
when $\sigma=1$, we obtain Sarsa, when $\sigma = 0$, we obtain Expected Sarsa and Tree-backup, and when $\sigma=1$ for every step of the backup except for the last, where $\sigma=0$, we obtain $n$-step Expected Sarsa. Thus, tuning the hyper-parameter $\sigma$ is not strictly necessary since it can be set to a fixed value in order to obtain one of the existing TD control algorithms.
Nevertheless, intermediate values of $\sigma$ between 0 and 1 create entirely new algorithms that exist somewhere between full sampling and pure expectation and that could result in better performance. 
Furthermore, $\sigma$ does not need to remain constant throughout every episode or even at every time step during an episode or continuing task. 
$\sigma$ could be varied dynamically as a function of time, of the current state, or of some measure of the learning progress.
In particular, $\sigma$ could also be varied as a function of the episode number, which we investigate in our experiments. 
There are potentially a variety of effective schemes for choosing and varying $\sigma$, and would be a subject for further research.

\section{Experiments}
\label{sec:experiments}
% In this section, we explore the performance of the different atomic multi-step algorithms that we have presented.
% First, we evaluate their performance on a prediction task which motivates the unification Sarsa and Expected Sarsa, and we also introduce the idea of dynamically varying $\sigma$ to take advantage of each algorithm's particular performance characteristics.
% %by showing that neither of the algorithms dominates over the other. 
% Then, on a gridworld navigation problem, we show that it is possible to improve performance with $Q(\sigma)$ by using an intermediate or dynamically varying value of $\sigma$, and by increasing the length of the backup. 
% Finally, we investigate the performance of $Q(\sigma)$ as an approximate solution method for a problem with a continuous state space.
%Additionally, we test the idea of dynamic $\sigma$ by implementing a $Q(\sigma$) algorithm with an initial $\sigma = 1$ that decreases by a factor of $0.95$ at the end of each episode.

\subsection{19-State Random Walk}
\label{sec:rwalk}
\begin{figure}[tb]
\begin{center}
\includegraphics[width=\linewidth]{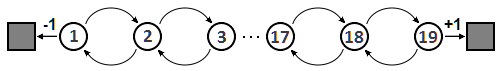}
\caption{The 19-state random walk MDP. The goal is to accurately estimate the value of each state under equiprobable random behavior.}
\label{fig:rw_mdp}
\end{center}
\end{figure}

The \textit{19-state random walk}, shown in Figure~\ref{fig:rw_mdp}, is a 1-dimensional environment where an agent randomly transitions to one of two neighboring states. There is a terminal state on each end of the environment, transitioning to one of them gives a reward of -1, and transitioning to the other gives a reward of 1. To compare algorithms that involve taking an expectation based on its policy, the task is formulated such that each state had two actions. Each action deterministically transitions to one of the two neighboring states, and the agent learns on-policy under an equiprobable random behavior policy. This differs from typical random walk setups where each state has one action that will randomly transition to either neighboring state~\cite{sutton98}, but the resulting state values are identical.

This environment was treated as a prediction task where a learning algorithm is to estimate a value function under its behavior policy. We conducted an experiment comparing various $Q(\sigma$) algorithm instances, assessing different multi-step backup lengths, step sizes, and degrees of sampling. The root-mean-square (RMS) error between its estimated value function and the analytically computed values was measured after each episode. Each $Q(\sigma)$ instance and parameter setting ran for 50 episodes and the results are averaged across 100 runs.

\begin{figure}[tb]
\begin{center}
\includegraphics[width=\linewidth]{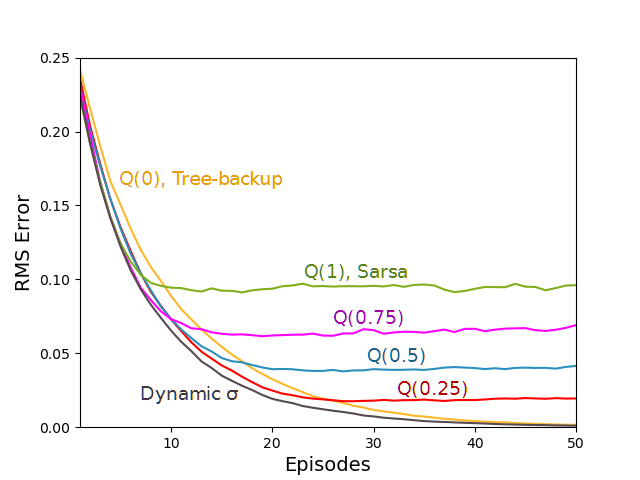}
\caption{19-state random walk results. The plot shows the performance of $Q(\sigma)$ in terms of RMS error in the value function. The results are an average of 100 runs, and the standard errors are all less than 0.006.
$Q(1)$ had the best initial performance, $Q(0)$ had the best asymptotic performance, and dynamic $\sigma$ outperformed all fixed values of $\sigma$.}
\label{fig:rw_results}
\end{center}
\end{figure}

Figure~\ref{fig:rw_results} shows the results with $n$ = 3 and $\alpha$ = 0.4, which was found to be representative of the best parameter setting for each instance of $Q(\sigma)$ on this task. Sarsa (full sampling) had better initial performance but poor asymptotic performance, Tree-backup (no sampling) had poor initial performance but better asymptotic performance, and intermediate degrees of sampling traded off between the initial and asymptotic performances.
This motivated the idea of dynamically decreasing $\sigma$ from 1 (full sampling) towards 0 (pure expectation) to take advantage of the initial performance of Sarsa, and the asymptotic performance of Tree-backup.
To accomplish this we decreased $\sigma$ by a factor of 0.95 after each episode.
$Q(\sigma)$ with a dynamically varying $\sigma$ outperformed all of the fixed degrees of sampling.

%which starts with full sampling and gradually shifts towards no sampling, 
%Additionally, we test the idea of dynamic $\sigma$ by implementing a $Q(\sigma$) algorithm with an initial $\sigma = 1$ that decreases by a factor of $0.95$ at the end of each episode.

\subsection{Stochastic Windy Gridworld}
\label{sec:windy}
% \begin{figure}[tb]
% \begin{center}
% \includegraphics[scale=0.3, trim={0 12cm 0 2cm}, clip]{windy.pdf}
% \caption{The windy gridworld as described by Sutton and Barto (1998). The start and goal states are denoted by \textit{S} and \textit{G} respectively. The numbers underneath each column denote the strength of the upward ``wind" experienced by the agent when it moves into that column.}
% \label{fig:windylayout}
% \end{center}
% \end{figure}
The \textit{windy gridworld} is a tabular navigation task in a standard gridworld which is described by Sutton and Barto~(\citeyear{sutton98}).
There is a start state and a goal state, and there are four possible moves: right, left, up, and down.
When the agent moves into one of the middle columns of the gridworld, it is affected by an upward ``wind'' which shifts the resultant next state upwards by a number of cells and varies from column to column.
% Figure \ref{fig:windylayout} shows the layout of the windy gridworld along with the strengths of the wind in each column.
If the agent is at the edge of the world and selects a move that would cause it to leave the grid, or would be pushed off the world by the wind, it is simply replaced in the nearest state at the edge of the world.
At each time step the agent receives a constant reward of -1 until the goal is reached.

\begin{figure}[tb]
\begin{center}
\includegraphics[width=\linewidth]{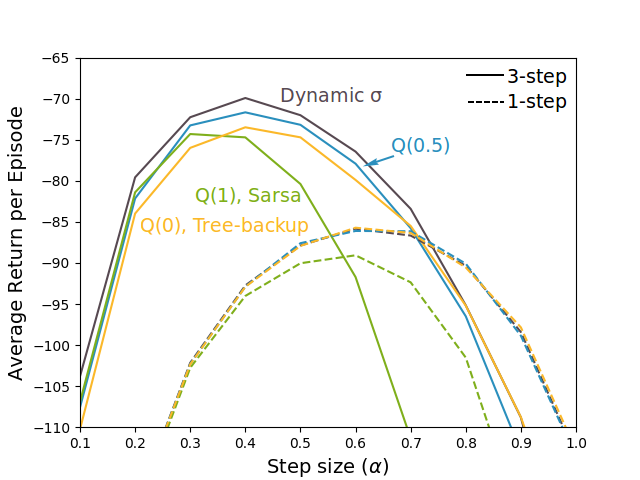}
\caption{Stochastic windy gridworld results. The plot shows the performance of $Q(\sigma$) in terms of the average return over 100 episodes as a function of the step size, $\alpha$, for various values of $\sigma$. The results are for selected $\alpha$ values, then are connected by straight lines, and are an average of 1000 runs. The standard errors are all less than 0.3 which is about a line width. 3-step algorithms performed better than their 1-step equivalents, and $Q(\sigma)$ with a dynamic $\sigma$ performed the best overall.}
\label{fig:stochasticwindyplots}
\end{center}
\end{figure}

A variation of the windy gridworld, called the \textit{stochastic windy gridworld}, is one where the results of choosing an action are not deterministic.
The layout, actions, and wind strengths are the same,
but at each time step, with a probability of 10\%, the next state that results from picking any action is determined at random from the 8 states currently surrounding the agent.

We conducted an experiment on the stochastic windy gridworld which consisted of 1000 runs of 100 episodes each to evaluate the performance of various instances of $Q(\sigma)$ with different parameter combinations.
All instances of the algorithms behaved and learned according to an $\epsilon$-greedy policy, with $\epsilon=0.1$.
As the performance measure, we compared the average return over the 100 episodes.
The results are summarized in Figure~\ref{fig:stochasticwindyplots}.

For all the values of $\sigma$ that we tested, choosing $n=3$ resulted in the greatest performance;
higher and lower values of $n$ decreased the performance.
Overall, $Q(\sigma$) with a dynamic $\sigma$ performed the best, while $\sigma=0.5$ was a close second.

\subsection{Mountain Cliff}
\begin{figure}[tb]
\begin{center}
\includegraphics[width=\linewidth]{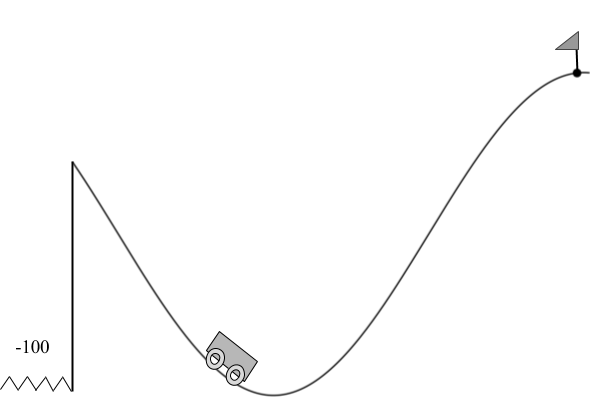}
\caption{The mountain cliff environment. The goal of the agent is to drive past the flag without falling off the cliff. The agent receives a reward of -1 at every time step, and falling off the cliff returns it to a random initial location in the valley with a reward of -100.}
\label{fig:mountainCliff}
\end{center}
\end{figure}
We implemented a variant of the classical episodic task, \textit{mountain car}, as described by Sutton and Barto~(\citeyear{sutton98}). For this implementation, the rewards, actions and goal remained the same. However, if the agent ever ventured past the top of the leftmost mountain, it would fall off a cliff, be rewarded -100 and returned to a random initial location in the valley between the two hills. We named this environment \textit{mountain cliff}. Both environments were tested and showed the same trend in the results. However, the results obtained in mountain cliff were more pronounced and thus were more suitable for demonstration purposes.

Because the state space is continuous, we approximated $q_\pi$ using \textit{tile coding} function approximation. Specifically, we used version 3 of Sutton's tile coding software (n.d.) with 8 tilings, an asymmetric offset by consecutive odd numbers, and each tile taking over 1/8 fraction of the feature space, which gives a resolution of approximately 1.6\%. 

For each algorithm, we conducted 500 independent runs of 500 episodes each. All training was done on-policy under an $\epsilon$-greedy policy with $\epsilon=0.1$ and $\gamma = 1$. We optimized for the average return after 500 episodes over different values of the step size parameter, $\alpha$, and the backup length, $n$. 
The results correspond to the best-performing parameter combination for each algorithm: $\alpha=1/6$ and $n=4$ for Sarsa;
$\alpha=1/6$ and $n=8$ for Tree-backup;
$\alpha=1/4$ and $n=4$ for $Q(0.5)$;
and $\alpha=1/7$ and $n=8$ for Dynamic $\sigma$.
We omit $n$-step Expected Sarsa in the results because its performance was not much different from $n$-step Sarsa's performance.

\begin{figure}[t]
\begin{center}
\includegraphics[width=\linewidth]{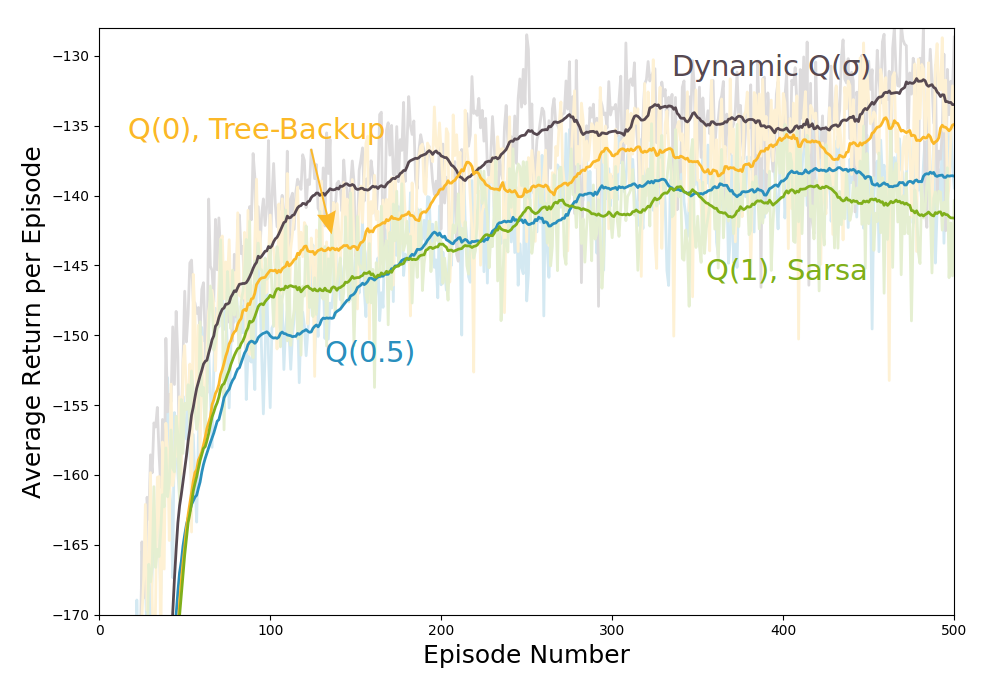}
\caption{Mountain cliff results. The plot shows the performance of each atomic multi-step algorithm in terms of the average return per episode. The dark lines show the results smoothed using a right-centered moving average with a window of 30 successive episodes, while the light lines show the un-smoothed results. $Q(\sigma)$ with dynamic $\sigma$ had the best performance among all the algorithms.}
\label{fig:mountainCliffResults}
\end{center}
\end{figure}

Figure \ref{fig:mountainCliffResults} shows the return per episode averaged over 500 runs. To smooth the results, we computed a right-centered moving average with a window of 30 successive episodes. Additionally, we added the average return per episode in a lighter tone to show the variance of each algorithm. As it can be observed, atomic multi-step Sarsa and $Q(0.5)$ had fairly similar performance. Among the atomic multi-step methods with static $\sigma$, Tree-backup had the best performance. Nonetheless, $Q(\sigma)$ with dynamic $\sigma$ outperformed all the algorithms that were using static $\sigma$.

In order to gain more insight into the nature of the results, we looked at the average return per episode after 50 (initial performance) and 500 (asymptotic performance) episodes for each algorithm. Additionally, a 95\% confidence interval was computed in order to validate the results.
% \footnote{The full results of this analysis are presented in the appendix.} 
After 50 episodes, $Q(0.5)$ had the best performance among the four algorithms with an average return per episode of -398.0; Dynamic $\sigma$ was a close second with an average return per episode of -406.3. On the other hand, after 500 episodes, Dynamic $\sigma$ managed to outperform all the other algorithms with an average return per episode of -163.7 followed by $Q(0.5)$ with an average return per episode of -167.9. $Q(1)$ (Sarsa) had the lowest performance with -447.3 average return per episode after 50 episodes and -173.2 after 500 episodes. These results contrast with Figure \ref{fig:mountainCliffResults} because the average is taken over all the previous episode instead of the preceding 30 episodes.
%such as in \ref{fig:mountainCliffResults}.

% we summarized the average return after 50 and 500 episodes in Table \ref{tbl:MountainCliff}. The standard error (SE), and lower (LB) and upper (UB) 95\% confidence interval bounds are provided to validate the significance of the results. All the summaries were calculated based on 500 runs.

%The average return after only 50 episodes could be interpreted as a measure of how fast the algorithm can learn, whereas the average return after 500 episodes shows how well an algorithm is capable of learning. As evidenced in the table, $Q(0.5)$ obtained the best performance during the first 50 episodes, while $Q(\sigma)$ with dynamic $\sigma$ was a close second. However, after 500 episodes, $Q(\sigma)$ with dynamic $\sigma$ managed to outperform all the other algorithms.

%%%%%%%%%%%%%%% Move to appendix?? %%%%%%%%%%%%%%%%%%%%%%%%%
% \begin{table}[t] 
% \caption{Summaries of the initial and final performance in terms of average return per episode for all the atomic multi-step algorithms in the mountain cliff environment. The standard error (SE), and lower (LB) and upper (UB) 95\% confidence interval bounds are provided to validate the results. 
% %$Q(0.5)$ had the best initial performance, whereas Dynamic $Q(\sigma)$ had the best final performance.
% }
% \label{tbl:MountainCliff}
% \begin{center}
% \begin{tabular}{lccccccccc}
% \toprule
% & \multicolumn{4}{c}{After 50 Episodes} &\hphantom{a}& \multicolumn{4}{c}{After 500 Episodes}\\
% \cmidrule{2-5}
% \cmidrule{7-10}
% Algorithm & Mean & SE & LB & UB &\hphantom{a}& Mean & SE & LB & UB\\
% \midrule
% $Q(1)$, Sarsa 		& -447.3		   & 1.15	& -449.2 		    & -445.4			&& -173.2		   	&  0.15 & -173.5 	 	   & -173.0	    	\\
% $Q(0)$, Tree-backup & -429.2 		   & 1.77	& -432.1 		    & -426.2 			&& -168.4		   	&  0.22 & -168.7 		   & -168.0			\\
% $Q(0.5)$  			& \textbf{-398.0}  & 1.11	& \textbf{-399.8}  	& \textbf{-396.1}	&& -167.9		   	&  0.15 & -168.2	      & -167.7	  		\\
% Dynamic $\sigma$ 	& -406.3 		   & 2.01	& -409.7 		    & -403.0 			&& \textbf{-163.7}  &  0.24 & \textbf{-164.1} & \textbf{-163.3}	\\
% \bottomrule
% \end{tabular}
% \end{center}
% \end{table}

\section{Discussion}
\label{sec:discussion}
From our experiments, it is evident that there is merit in unifying the space of algorithms with $Q(\sigma)$. In prediction tasks, such as the 19-state random walk, varying the degree of sampling results in a trade-off between initial and asymptotic performance. In control tasks, such as the stochastic windy gridworld, intermediate degrees of sampling are capable of achieving a higher per-episode average return than either extreme, depending on the number of elapsed episodes.

These findings also extend to tasks with continuous state spaces, such as the mountain cliff.
% As evidenced by the results in Table \ref{tbl:MountainCliff}, 
Intermediate values of $\sigma$ allow for a higher initial performance, whereas small values of $\sigma$ allow for a better asymptotic performance. As shown in Figure \ref{fig:mountainCliffResults}, $Q(\sigma)$ with dynamic $\sigma$ is able to exploit these two benefits by adjusting $\sigma$ over time.

Moreover, our experiments in the stochastic windy gridworld task demonstrated that it is possible to improve performance by choosing a higher value of the backup length parameter, $n$.
Varying $n$ controls a bias-variance trade-off by adjusting how many rewards are included in the backup before bootstrapping, similar to the parameter $\lambda$ in the TD($\lambda$) algorithm. The parameter $\sigma$ also has a bias-variance trade-off interpretation, as the Tree-backup algorithm decays the weighting of future rewards based on the stochasticity in the policy (and is therefore more biased). The length parameter $n$ controls the bias-variance trade-off in the direction of the trajectory taken, while the parameter $\sigma$ manages it by controlling the bootstrapping in the direction of actions not taken. 
% Added to address reviewer 3's comments (see third item in the checklist)
A qualitative result that illustrates the bias-variance trade-off induced by the parameter $\sigma$ can be observed in the 19-State Random Walk experiment. A large value of $\sigma$ results in lower bias at the beginning of training and a lower RMS error as a consequence. However, as the bias of the return decreases in the asymptote, the low variance inherent to small values of $\sigma$ result in more accurate estimates of the action-value function.

% When comparing algorithms with a backup length greater than one, we noticed that the Tree-backup algorithm experienced a shorter \textit{effective backup length}. 
% For example, with no discounting ($\gamma$ = 1), $n$-step Sarsa's effective backup length is equal to its backup length $n$ because the TD error at each step is equally weighted. However, the effective backup length of Tree-backup can be less than $n$ depending on the stochasticity of $\pi$; this is a direct result from the product term in (\ref{eqn:treeBackup}), giving less weight to the TD errors of later steps.
% Thus, for some algorithms the length of the backup, $n$, and the effective backup length are not equal. 
% We did not explore this idea any further, but it could be subject for new research.

% - todo: briefly mention/summarize effective length of tree-backup being shorter
% - explain how sampling can lead to better updates if the sampled action has a reasonably accurate value

\section{Conclusions}
\label{sec:conclusion}
In this paper we studied $Q(\sigma)$, which is a unifying algorithm for multi-step TD control methods.
$Q(\sigma)$, through the use of the sampling parameter $\sigma$, allows for continuous variation between updating based on full sampling and updating based on pure expectation.
Our results on prediction and control problems showed that an intermediate fixed degree of sampling can outperform the methods that exist at the extremes (Sarsa and Tree-backup). 
In addition, we presented simple way of dynamically adjusting $\sigma$ which outperformed any fixed degree of sampling.

Our presentation of $Q(\sigma)$ was limited to the atomic multi-step case without eligibility traces, we only conducted experiments on on-policy problems, and we only investigated one simple method for dynamically varying $\sigma$.
This leaves open several avenues for future research.
First, $Q(\sigma)$ could be extended to use eligibility traces and compound backups.
Second, the performance of $Q(\sigma)$ could be evaluated on off-policy problems.
Third, other schemes for dynamically varying $\sigma$ could be investigated
-- perhaps as a function of state, the recently observed rewards, or some measure of the learning progress.

\section{Acknowledgments}
The authors thank Vincent Zhang, Harm van Seijen, Doina Precup, and Pierre-luc Bacon for insights and discussions contributing to the results presented in this paper, and the entire Reinforcement Learning and Artificial Intelligence research group for providing the environment to nurture and support this research. We gratefully acknowledge funding from Alberta Innovates -- Technology Futures, Google Deepmind, and from the Natural Sciences and Engineering Research Council of Canada.

\section{Appendix}
\subsection{Proof of Theorem 1}

Let $\mathcal{X} = \mathcal{S} \times \mathcal{A}$, $X_t = (S_t, A_t) \in \mathcal{X}$, $\bar{R}_t = \mathbb{E}\{R_t\}$, and $Q*$ be the optimal action-value function defined as
\begin{align}
Q^*(S_t, A_t) = \bar{R}_{t+1} + \gamma \mathbb{E}\{ \max_a Q^*(S_{t+1},a) \}.
\label{eqn:optimalav}
\end{align}
We define a new stochastic process $(\alpha_t, \Delta_t, F_t)_{t \geq 0}$ by subtracting $Q^*(X_t)$ from both sides of equation (\ref{eqn:1stepqs}) 
\begin{align*}
\Delta_{t+1}(X_{t}) = (1-\alpha_t(X_t)) \Delta_{t}(X_{t}) - \alpha_t(X_t)F_t(X_t), 
\end{align*}
and letting $\alpha_t \in (0,1]$, $\Delta_t(X_t) = Q_t(X_t) - Q^*(X_t)$, and $F_t = R_{t+1} + \gamma [\sigma_{t+1} Q_{t}(X_{t+1}) + (1-\sigma_{t+1})V_{t+1}] -  Q^*(X_t)$. Additionally, let $P_t$ be a sequence of increasing $\sigma$-fields representing the history such that $\alpha_0$ and $\Delta_0$ are $P_0$-measurable and $\alpha_t$, $\Delta_t$, and $F_{t-1}$ are $P_t$-measurable for $t \geq 1$.

Proving that $\Delta_t$ converges to $0$ as $t \rightarrow \infty$ is equivalent to showing that $Q_t$ converges to $Q^*$ as $t \rightarrow \infty$. Consequently, the proof is equivalent to showing that the conditions of lemma 1 from Singh et al. (\citeyear{Singh2000}) are satisfied for $\Delta_t$. 

Conditions one, two, and three of the lemma are satisfied by the corresponding assumptions of the theorem. Hence, we only need to show that $||\mathbb{E}\{ F_t | P_t \}|| \leq k ||\Delta_t|| + C_t$ where $||.||$ is the maximum norm, $k \in [0,1)$, and $C_t$ goes to $0$ with probability 1. By adding and subtracting $\max_a Q_t(S_t,a)$, using the definition of $Q^*$ and the triangle inequality, we can show that
\begin{align}
& ||\mathbb{E}\{F_t | P_t\}|| \nonumber \\  
& \quad \leq || \mathbb{E}\{R_{t+1} + \gamma \max_a Q_{t}(S_{t+1},a) - Q^*(S_t,A_t)\}|| \nonumber \\
& \qquad + \gamma || \mathbb{E}\{\sigma_{t+1} Q_t(s_{t+1},a_{t+1}) + (1-\sigma_{t+1})V_{t+1} \nonumber \\
& \qquad \qquad - \max_a Q_{t}(s_{t+1},a)\}|| \nonumber \\
& \quad = \gamma || \mathbb{E}\{\max_a Q_t(S_{t+1},a) - \max_b Q^*(S_{t+1},b)\}|| + C_t \nonumber \\
& \quad \leq \gamma \max_s | \max_a Q_t(s,a) - \max_b Q^*(s,b)| + C_t \nonumber \\
& \quad \leq \gamma \max_s \max_a | Q_t(s,a) - Q^*(s,a)| + C_t \nonumber \\
& \quad = \gamma ||\Delta_t|| + C_t. \nonumber
\end{align}

Note that if the policy is greedy and $\sigma_{t+1} \in [0,1]$, then $\sigma_{t+1}Q_t(S_{t+1},A_{t+1}) + (1-\sigma_{t+1})V_{t+1} = \max_a Q_t(S_{t+1},a)$. Therefore, $C_t$ goes to $0$ as the policy becomes greedy in the limit. Consequently, condition 3 of lemma 1 from Sing et al. (\citeyear{Singh2000}) is satisfied. Therefore, $\Delta_t$ converges to 0 w.p. 1, which implies that $Q_t$ converges to $Q^*$ w.p. 1. \qed

\bibliography{references}
\bibliographystyle{aaai}
\end{document}